\def\arxivversion{}
\documentclass[10pt,a4paper]{article}

\usepackage[utf8]{inputenc}
\usepackage[T1]{fontenc}
\usepackage[english]{babel}

\usepackage{amsmath,amssymb}
\usepackage{graphicx}
\usepackage{multicol}

\usepackage{newtxtext}
\usepackage{newtxmath}
\usepackage{booktabs}
\usepackage{xcolor}
\usepackage[normalem]{ulem}
\usepackage{soul}
\usepackage[hidelinks]{hyperref}
\usepackage{tikz}
\usetikzlibrary{positioning,arrows.meta,shapes.geometric,fit,backgrounds,calc}

\usepackage[a4paper,left=2cm,right=2cm,top=2.54cm,bottom=3.76cm]{geometry}

\usepackage{microtype}
\usepackage{caption}
\usepackage{subcaption}
\usepackage{titlesec}
\titleformat{\section}
  {\normalfont\bfseries\MakeUppercase}{\new{\thesection.}}{0.5em}{}
\titleformat{\subsection}
  {\normalfont\bfseries}{\new{\thesubsection}}{0.5em}{}
\titleformat{\subsubsection}
  {\normalfont\itshape}{\new{\thesubsubsection}}{0.5em}{}
\titlespacing*{\section}      {0pt}{0.6\baselineskip}{0.3\baselineskip}
\titlespacing*{\subsection}   {0pt}{0.45\baselineskip}{0.2\baselineskip}
\titlespacing*{\subsubsection}{0pt}{0.35\baselineskip}{0.15\baselineskip}
\titleformat{\paragraph}[runin]{\normalfont\bfseries}{}{0pt}{}[.]
\titlespacing*{\paragraph}{0pt}{0.2\baselineskip}{0.5em}

\usepackage{fancyhdr}
\newcommand{\obpdcfooter}{\ifdefined\arxivversion\else OBPDC 2026\fi}
\fancypagestyle{plain}{%
  \fancyhf{}%
  \fancyfoot[C]{\obpdcfooter}%
}

\usepackage[numbers,sort&compress]{natbib}

\graphicspath{{figures/}}

\newif\ifmarkup
\ifdefined\diffversion \markuptrue \else \markupfalse \fi
\ifmarkup
  \newcommand{\new}[1]{{\color{red}#1}}
  \newcommand{\del}[1]{{\color{red}\sout{#1}}}
  \sethlcolor{yellow}
  \newcommand{\newy}[1]{\hl{#1}}
  \newcommand{\dely}[1]{\colorbox{yellow}{\sout{#1}}}
  \newcommand{\figupdated}{ \hl{(updated figure)}}
\else
  \newcommand{\newy}[1]{#1}
  \newcommand{\dely}[1]{\unskip}
  \newcommand{\figupdated}{}
  \newcommand{\new}[1]{#1}
  \newcommand{\del}[1]{\unskip}
\fi
\newcommand{\secref}[1]{\new{\ref{#1}}}
\colorlet{RED}{red}
\ifmarkup
  \newcommand{\todo}[2]{\textcolor{red}{\textbf{[TODO #1: #2]}}}
\else
  \newcommand{\todo}[2]{\ignorespaces}
\fi

\renewenvironment{abstract}{%
  \par\vspace{0.2\baselineskip}%
  {\normalfont\bfseries ABSTRACT}\par\vspace{0.3\baselineskip}%
}{\par\vspace{0.4\baselineskip}}

\title{\fontsize{12}{14}\selectfont\bfseries A GENERIC ON-BOARD AI PROCESSING CHAIN
FOR EARTH OBSERVATION: FROM APPLICATION DEVELOPMENT TO EMBEDDED DEPLOYMENT AND
MISSION-LEVEL ORCHESTRATION\ifdefined\arxivversion$^{*}$\fi}

\author{%
\normalsize\bfseries
Michel\new{-}Fran\new{\c{c}}ois Foulon\textsuperscript{(1)}, Liz Ramos Medina\textsuperscript{(1)},
Thomas Goudemant\textsuperscript{(\new{1,}2)}, Clotilde Szywala\textsuperscript{(2)},
Virgile Robles\textsuperscript{(3)},\\
\normalsize\bfseries
Cl\new{\'e}ment Petitjean\textsuperscript{(1)}, Ad\`ele Hankache\textsuperscript{(1)},
Omar Hlimi\textsuperscript{(2)}\\[0.4ex]
\normalfont\normalsize\bfseries\itshape
\textsuperscript{(1)}Thales Alenia Space\new{, France} \quad
\textsuperscript{(2)}IRT Saint Exup\'ery\new{, France} \quad
\textsuperscript{(3)}Tarides\new{, France}\\[0.2ex]
\normalfont\normalsize\itshape Emails: michel-francois.foulon@thalesaleniaspace.com,\\
\normalfont\normalsize\itshape thomas.goudemant@irt-saintexupery.com, virgile@tarides.com
}

\date{}

\begin{document}
\maketitle
\vspace{-2\baselineskip}
\thispagestyle{plain}
\ifdefined\arxivversion
  {\renewcommand{\thefootnote}{*}%
  \footnotetext{Accepted at the 10th On-Board Payload Data Compression Workshop
  (OBPDC 2026), Barcelona, Spain, 12--14 October 2026.}}
\fi

\begin{abstract}
\new{Earth observation satellites acquire more data than they can downlink.
Extracting information on board with artificial intelligence (AI) turns the
satellite into a data-reduction stage that downlinks compact products, such as
masks or detections, instead of full scenes; deploying it in a mission,
however, involves the whole processing chain and not only the neural network.
We present a generic on-board AI processing chain along three axes, with wildfire segmentation as the worked
example among ship detection and cloud segmentation.
(i)~The AI model is tuned (spectral-band selection, input normalization),
quantized to 8-bit integers (int8) and ported to an AMD Versal VCK190, with its excellent
segmentation performance preserved on the board.
(ii)~A modular tiler--inference--mosaicer workflow, reused unchanged across
applications, brings the tile-level model to scene level, turning a full
acquisition into a scene-level mask or detection list.
(iii)~Each step is packaged as a unikernel and orchestrated on board by
ORCHIDE, a cloud-native orchestrator.
The chain is characterized end to end, from the standalone execution on the
target to its execution under orchestration with parallel workflows on a
representative test bench. Beyond AI inference on individual tiles, this
approach aims at increasing the maturity of on-board AI deployment by
addressing a complete processing chain, from the acquired scene to the
mission-level execution.}
\end{abstract}

\section{Introduction}
\label{sec:intro}

\new{Earth Observation (EO) satellites acquire ever-growing data volumes, while
downlink capacity, on-board mass memory and ground-processing latency remain
limiting factors~\cite{duggan2025survey}. Moving part of the processing on
board turns the satellite into a data-reduction stage: instead of downlinking
complete scenes with uniform priority, it extracts and transmits only the
actionable information (detections, masks, selected regions), which reduces the
data according to its content and complements conventional
compression~\cite{zhang2022intelligent}. Advances in embedded}
Artificial Intelligence (AI) and heterogeneous computing have enabled the
deployment of onboard processing chains capable of executing complex algorithms
directly on satellite hardware. However, the integration of such AI-driven
workflows introduces new complexities, including resource management, task
orchestration, and hardware acceleration, which must be efficiently coordinated
to ensure optimal performance.

Recent advancements in onboard satellite processing have demonstrated the
feasibility of deploying AI-driven applications for real-time environmental
monitoring: onboard anomaly detection for marine protection and ship detection\del{,}
leveraged embedded AI to reduce reliance on ground
stations~\cite{104goudemant2024anomaly80531,goudemant2022ships}. Similarly,
in-orbit AI-based cloud segmentation \del{on} has been validated on OPS-SAT,
showcasing efficient image processing via System-on-Module (SoM)
architectures~\cite{feresin2021opssat,kervennic2023edhpc}. Extending these
efforts, multi-spectral wildfire detection models further emphasized the role
of onboard AI in disaster monitoring~\cite{farhat2022wildfire}.
\new{In orbit, $\Phi$sat-1 discards cloudy images on
board~\cite{giuffrida2022phisat1} and $\Phi$sat-2 embeds a complete image
processing chain~\cite{melega2023phisat2}; cloud screening ahead of on-board
compression likewise avoids transmitting cloudy
pixels~\cite{cilia2020cloud}. Object detection has been studied on board
high-resolution imagery~\cite{shen2023onboard}, including a YOLOX
detector~\cite{ge2021yolox} on an on-orbit FPGA~\cite{wang2022yoloxfpga}, and
the Versal architecture targeted here has been evaluated for edge computing in
space~\cite{perryman2023versal}; a companion paper of this workshop discusses
the data reduction that on-board vessel detection brings to the
downlink~\cite{goudemant2026maritime}.}

In a real mission, such processing chains must be executed on the hardware for
each acquisition of sensor data. Orchestrators have proven necessary to match
these increasingly demanding in-orbit applications with the constrained
\new{resources} (CPU, RAM, storage) of satellites while enforcing the tight
schedule of a mission. ORCHIDE \newy{(Orchestration of Reliable Computing on
Heterogeneous Infrastructures Deployed at the Edge)}~\cite{hankache2025orchestrator} is such an
orchestrator aiming to adapt a cloud-native orchestrator (Argo/k3s) to a
spatial context while making no assumptions about the safety of the
orchestrated programs, by using unikernels\new{, i.e.\ applications packaged
with only the operating-system services they need into small isolated virtual
machines~\cite{madhavapeddy2013unikernels,kuenzer2021unikraft,weisz2026orchide}.
Our team also develops a ground/on-board demonstrator of a reactive loop for
EO missions, going from the detection of an event to the triggering of an
action~\cite{francesconi2025reactive}; the work presented here provides an
orchestration solution for its on-board AI part}.

\new{These works mostly assess individual models on tiles, or the orchestrator
alone; the generic chain around the model, from the acquired scene to its
execution under mission-level orchestration, is rarely characterized as a
whole.} In this paper, we present a generic onboard AI processing chain designed for EO
missions, covering the entire pipeline from application development to embedded
deployment and mission-level orchestration. Our approach leverages modular
workflows (i.e.\ chains of processing steps) and hardware-aware
characterization to enable seamless execution of AI models---such as ship
detection, wildfire segmentation and cloud segmentation---on
resource-constrained satellite platforms\new{; other on-board use cases, such as
marine anomaly detection~\cite{104goudemant2024anomaly80531}\new{, with first
in-orbit results~\cite{goudemant2025inorbit},} or the assessment
of destroyed buildings from bi-temporal
images~\cite{goudemant2026optimizing}, fit the same chain}. Each step of the
processing chain (tiling, inference and aggregation of results) is designed to
be a reusable and composable standalone program.

\new{The paper is organized as follows. Section~\secref{sec:method} describes the chain along its three
axes: the AI model, from data and model tuning to its quantization and porting
to the Versal VCK190, with wildfire segmentation as the worked example
(Section~\secref{sec:method-ai}); the generic tiler--inference--mosaicer
workflow that brings it to scene level (Section~\secref{sec:method-workflow}); and
its integration in the ORCHIDE on-board orchestrator, together with the test
bench used for characterization (Section~\secref{sec:method-orch}).
Section~\secref{sec:experiments} reports results at two levels: for the
standalone chain on the VCK190, the impact of band selection, input
normalization and 8-bit integer (int8) quantization on segmentation quality, and the latency
and power of each stage (Section~\secref{sec:exp-algo}); at workflow level, the overhead of executing the same chain as unikernels within ORCHIDE
and the gains brought by running several workflows in parallel
(Section~\secref{sec:exp-orch}). Section~\secref{sec:conclusion} summarizes the
results, their limitations and the perspectives for mission-level operation.}
\del{In section 4.3, this paper explores the feasibility and performance
consequences of embedding applications in an orchestration framework. This
evaluation illustrates the feasibility of orchestration pipelines on
satellites, and where there is room for improvement when embedding cloud-native
technology into constrained environments.}

\section{Method}
\label{sec:method}

The processing chain is organized along three axes: (i) the on-board AI core,
which runs a per-tile detection or segmentation model on the Versal VCK190;
(ii) the workflow that surrounds it, turning a full satellite scene into tiles
and re-assembling per-tile products into a scene-level result; and (iii) the
mission layer, which schedules and orchestrates these workflows on board. This
section describes the data, the models and the target for each axis;
quantitative results are reported in Section~\secref{sec:experiments}. Among the
three supported applications---ship detection, wildfire segmentation and cloud
segmentation---we use wildfire segmentation as the worked example throughout;
ship detection (a YOLOX detector\new{~\cite{ge2021yolox}}) and cloud
segmentation (a U-Net model) share the same chain and the same pre-processing
and quantization strategy, and are only referred to where relevant.

\subsection{\new{On-Board AI Model}: The Wildfire Segmentation Use Case}
\label{sec:method-ai}

\subsubsection{\new{Data}: Acquisition and Annotation}
\label{sec:method-data}
The wildfire dataset is built from Sentinel-2 L1C products downloaded from the
Copernicus ecosystem. We select 90 scenes covering active-fire events over
varied backgrounds, with the 13 spectral bands resampled to a common
20\,m/pixel grid and stored as multi-band TIFF images. Fire-front annotations
are produced following the multispectral wildfire-detection methodology of
Farhat et al.~\cite{farhat2022wildfire}, which exploits the strong short-wave
infrared (SWIR) response of active fires; each scene is paired with a binary
TIFF mask (fire = 1 / non-fire = 0).

Each scene is cut into $1{,}024\times1{,}024$ tiles with a 512-pixel stride,
matching the fixed tile size processed by the on-board AI core. The dataset is
split into training, validation and test sets at scene level: all tiles of a
given scene are assigned to a single set, so that no scene contributes tiles to
more than one set. This prevents spatial leakage between training and
evaluation and yields a realistic estimate of generalization to unseen scenes.
The resulting split contains 2,748 training tiles (${\sim}53\,\%$), 1,214
validation tiles (${\sim}24\,\%$) and 1,174 test tiles (${\sim}23\,\%$).

\subsubsection{Model and Spectral-Band Selection}
\label{sec:method-model}
We use a U-Net~\cite{ronneberger2015unet} for the segmentation task. Its
architecture is deliberately classical: a five-level convolutional encoder that
extracts increasingly abstract features, a symmetric decoder that restores full
resolution, skip connections that recover the spatial detail lost to pooling,
and a final $1\times1$ convolution producing a per-pixel score in $[0,1]$,
thresholded (e.g.\ at 0.5) to obtain a binary fire mask. This model is a good
fit for the target for three reasons: it is effective on the segmentation task,
it is lightweight (0.15\,M trainable parameters, ${\approx}9.4$\,GFLOPs per
$1{,}024\times1{,}024$ inference), and it is portable---its layers map cleanly
onto the operators supported by the embedded toolchain, which keeps porting and
debugging simple.

A Sentinel-2 tile carries 13 bands, but feeding all of them to the model is
neither necessary nor optimal. We therefore study which bands are actually
required to detect fire fronts, comparing three input configurations: all 13
bands (baseline), the three visible bands (B02, B03, B04), and a three-band
SWIR-oriented set (B02, B11, B12, i.e.\ blue, SWIR1, SWIR2). Based on the
ablation results reported in Section~\secref{sec:exp-algo-perf}, we retain the
B02, B11, B12 configuration.

\subsubsection{\new{Porting to the VCK190: Quantization and Input Normalization}}
\label{sec:method-quant}
The AI core is deployed on the AMD Versal VCK190, whose Deep-learning
Processing Unit (DPU) executes the model in 8-bit integer (int8) arithmetic.
Quantization is performed with the Vitis AI Post-Training Quantization (PTQ)
flow~\cite{amd_vitisai}, which calibrates \new{per-tensor} scale factors so
that activations and weights are represented in int8; the model is then
compiled to an xmodel for the DPU. Since the accuracy preserved by PTQ turns
out to be more than adequate (see Section~\secref{sec:exp-algo-perf}), we did not
pursue Quantization-Aware Training (QAT), which would have added a training
cost for a marginal expected gain.

The DPU input tensor is likewise expected in int8, which raises the question of
how to map the \new{16-bit unsigned integer (uint16)} Sentinel-2 pixels onto that range. The pre-processing
must satisfy a hard embedded constraint: it should avoid per-image,
data-dependent computations\new{, which are costly and make the on-board
processing time scene-dependent}. We considered three families of input
normalization: (i) percentile normalization, which clips each band to its
1st/99th percentiles before scaling---accurate but requiring a per-band
histogram on every tile, i.e.\ a costly, scene-dependent computation on board;
(ii) per-band standardization (centering and reducing each band), which
likewise depends on scene statistics; and (iii) fixed scaling by a constant. As
explained in Section~\secref{sec:exp-algo-perf}, we adopt the last option: uint16
pixels are divided by the constant 65,535, which the PTQ input scale then maps
to int8. This is a single elementwise operation with no histogram and no scene
statistics, and it is compatible with all three applications, enabling a
single, common, effective and computationally efficient pre-processing
strategy.

\subsection{\new{From Tiles to Scenes: The} Inference Workflow}
\label{sec:method-workflow}

\begin{figure}[t]
  \centering
  \includegraphics[width=0.97\textwidth]{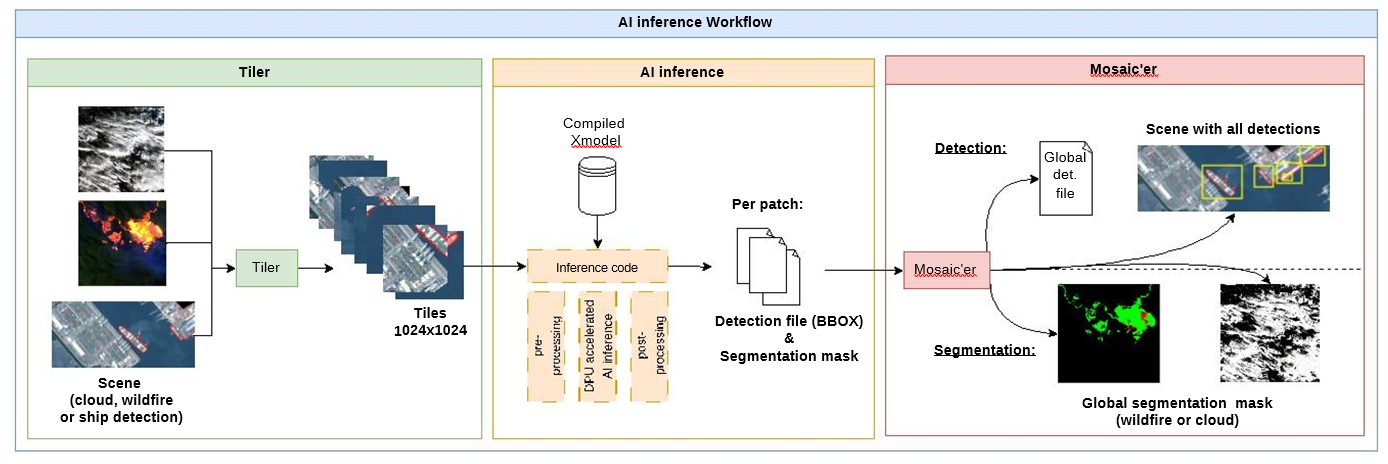}
  \caption{End-to-end inference workflow: the tiler cuts a full scene into
  $1{,}024\times1{,}024$ tiles, the AI inference stage includes per-tile
  pre-processing, DPU-accelerated inference and post-processing on the VCK190,
  and the mosaicer re-assembles the per-tile products. \new{Maxar Products
  Standard (2A) Imagery \copyright 2017--2024 Maxar Technologies.}}
  \label{fig:workflow}
\end{figure}

The AI core does not operate on full scenes but on fixed-size tiles, so it is
inserted between two lightweight stages: a tiler that feeds it and a mosaicer
that collects its outputs (Fig.~\ref{fig:workflow}). This generic, modular
chain is reused unchanged across the three applications; only the embedded
model and its band selection differ. Each stage is packaged as a standalone
application, i.e.\ a minimal, composable and reusable execution unit.

\subsubsection{Tiler}
\label{sec:method-tiler}
The tiler turns a large satellite scene (a stacked multi-band Sentinel-2 scene
for wildfire and cloud; a pansharpened scene for ships) into a set of square
$1{,}024\times1{,}024$ tiles. Two elementary pre-processing steps---the band
selection and the \new{input rescaling}\del{uint16$\to$int8 conversion} (i.e.\ the division by 65,535 of
Section~\secref{sec:method-quant})---can be placed either here, in the tiler, or
in the AI inference core. Keeping the tiler agnostic (tiling only, raw uint16
tiles) produces tiles that any downstream model can exploit while selecting its
own bands, which keeps the chain generic. Conversely, when several models share
the same bands, performing the selection and \new{rescaling}\del{conversion} once in the tiler
avoids repeating a common pre-processing per model. The chain supports both
placements; the choice is a mission-integration decision rather than an
algorithmic one.

\subsubsection{Mosaicer}
\label{sec:method-mosaicer}
The mosaicer aggregates per-tile products into a single scene-level result. For
segmentation (wildfire, cloud) it concatenates the per-tile masks back into one
full-scene mask. For detection (ships) it merges the per-tile bounding-box files
into a single scene-level file, shifting each tile's local coordinates by the
pixel offset of the tile's top-left corner so that all detections are expressed
in scene coordinates.
\new{These scene-level products are what is downlinked, and this is where the
chain reduces the data: a binary wildfire or cloud mask can be encoded on 1 bit
per pixel, against 16 bits per band and per pixel for the input scene (208 bits
per pixel for the 13 Sentinel-2 bands), and a ship detection reduces to a few
bytes per bounding box. How such products translate into downlink modes and
volumes is studied for vessels in a companion
paper~\cite{goudemant2026maritime}.}

\subsection{Mission-Level Integration \new{and On-Board} Orchestration}
\label{sec:method-orch}

\subsubsection{\new{Processing Workflows in ORCHIDE}}
\label{sec:method-orch-workflows}
Integration at mission level requires considering the whole processing chain
from sensor acquisitions to inference outputs while managing on board
resources. The ORCHIDE orchestrator is made responsible \new{for}\del{of} both acquisition
management (following a mission plan) and workflow scheduling taking into
account hardware resources. In this paper, acquisitions are considered out of
scope and the focus is put on the formalization of processing chains through
dedicated ``workflows'' and their execution.

A workflow is a composition of elementary processing steps working on
identified inputs and producing predictable outputs. Within a workflow, each
elementary processing step must be fully executed before moving to the next
one. Concretely in ORCHIDE, each elementary processing step is instantiated as
a Kubernetes pod. The maximum number of workflows that can be orchestrated
concurrently can be configured in the orchestrator. ORCHIDE then matches \new{pods} and
their resource requirements (defined in the workflow specification) to
available computation resources, as illustrated \new{in}\del{on} Fig.~\ref{fig:pods}. While
the allocation of CPU time and memory behaves as expected, the DPU required by
the AI core is indivisible, meaning that while multiple inference steps may be
executing at the same time, the actual inference requests will be queued and
treated sequentially. Section~\secref{sec:exp-orch} explores the actual
performance gains observed when executing multiple workflows in parallel.
At a mission level, a mission plan provides instructions to ORCHIDE for (1)
defining the start and end time of acquisitions and (2) specifying which
workflows must run for each acquisition. For proper time and hardware resources
\new{allocation}\del{affectation} for workflow operations (2) on a given volume of acquisitions (1),
an estimation of end-to-end inference workflow performances is necessary on
representative hardware targets, which is \new{the} purpose of the experiments described
in Section~\secref{sec:exp-orch} that run on the \new{test bench described
below}\del{above test bench}.

\subsubsection{\new{Test Bench for} Hardware-Target Characterization}
\label{sec:method-bench}
\begin{figure}[t]
  \centering
  \begin{minipage}[b]{0.47\textwidth}
    \centering
    \includegraphics[width=\linewidth]{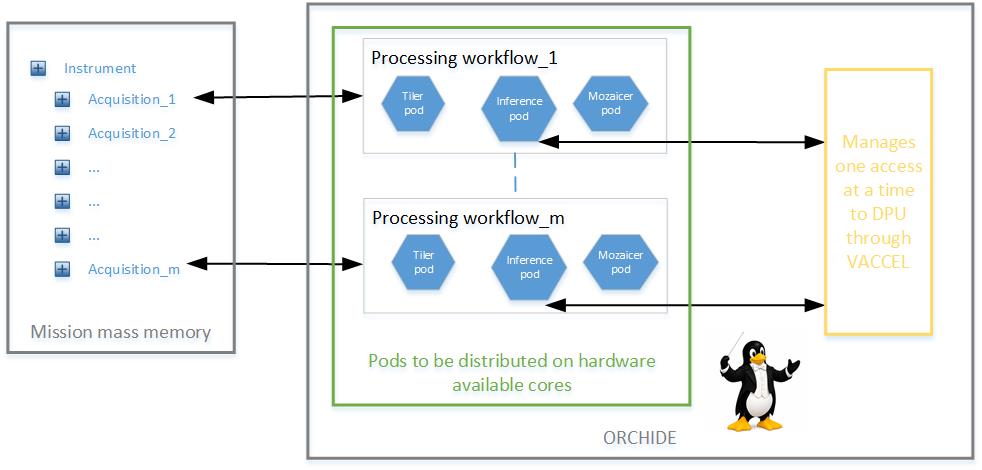}
    \caption{Pods association to end to end inference process.}
    \label{fig:pods}
  \end{minipage}\hfill
  \begin{minipage}[b]{0.5\textwidth}
    \centering
    \includegraphics[width=\linewidth]{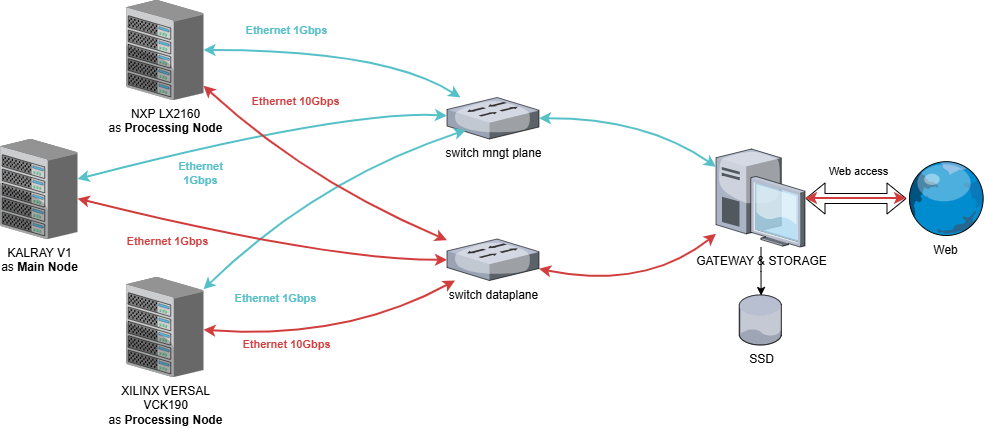}
    \caption{ORCHIDE test \new{bench} for workflow characterization.\figupdated}
    \label{fig:testbench}
  \end{minipage}
\end{figure}

We set up a test bench with the topology illustrated in
Fig.~\ref{fig:testbench} to characterize the performance of an end-to-end use
case across heterogeneous hardware using the orchestrator, replicating previous
ORCHIDE experiments~\cite{weisz2026orchide}. It consists of a management plane
(blue links) performing pod loading and a data plane (red links) responsible for
exchange of acquisition and intermediate results. The orchestrator function is
carried out by a KALRAY node while a NXP LX2160 node and Xilinx Versal VCK190
nodes are used as processing nodes on which actual steps are executed.

\section{Experiments and Results}
\label{sec:experiments}

Section~\secref{sec:exp-algo} measures the ideal performance of each individual
processing step, while Section~\secref{sec:exp-orch} explores the impact of
embedding these applications in an end-to-end orchestrator such as ORCHIDE.

\subsection{\new{Standalone Performance on the VCK190: Algorithm and Hardware}}
\label{sec:exp-algo}

\subsubsection{Experimental Setup}
\label{sec:exp-setup}
All embedded measurements at algorithm level are taken on an AMD Versal VCK190
board. Models are trained in floating point on a server, quantized to int8 and
compiled with Vitis AI 3.0~\cite{amd_vitisai}, then deployed on the board's
DPU. For wildfire we report results on the full test set of 1,174
$1{,}024\times1{,}024$ tiles (Section~\secref{sec:method-data}). Algorithmic
quality is measured with the F1-score of the fire class; hardware performance
is measured as the latency of each processing stage for a batch of six tiles,
from which we derive throughput in megapixels per second (Mpix/s).

\subsubsection{Algorithmic Performance}
\label{sec:exp-algo-perf}

\paragraph{Spectral-band selection}
Table~\ref{tab:bands} compares the three input configurations of
Section~\secref{sec:method-model} for the floating-point model. For this ablation,
the three configurations are trained under identical training conditions on a
reduced dataset of 183 patches extracted from 12 scenes. The resulting
floating-point models are then evaluated and compared in terms of algorithmic
performance. Results show that the visible-only set (B02, B03, B04) collapses,
as visible bands do not carry enough information to localize fire fronts. In
contrast, the SWIR-oriented set (B02, B11, B12) captures the thermal-emission
signature of active fires highlighted by Farhat et
al.~\cite{farhat2022wildfire}. It matches the 13-band baseline within a
fraction of a point in F1 while reducing the input tensor from 13 to 3 channels
and showing less overfitting. We therefore retain B02, B11, B12 for all
subsequent experiments.
\paragraph{Input normalization and quantization}
The pre-processing has a first-order effect on how well accuracy survives int8
quantization. We trained the model under identical conditions on the full
training set with representatives of the three normalization families of
Section~\secref{sec:method-quant} and measured the F1-score of both the
floating-point and the quantized model on the full test set
(Table~\ref{tab:norm}). Percentile clipping and per-band centering are either
fragile through quantization (the latter collapses to near-zero F1) or require
scene-dependent computation on board. Fixed scaling by 65,535 provides the best
trade-off between algorithmic performance and on-board computational cost: it
reaches 95.10\,\% F1 in floating point and 94.68\,\% after quantization, a drop
of only 0.42 percentage points, while requiring only a single elementwise
division, with no per-image histogram or scene statistics. Experiments on the
other two applications led to the same conclusion, supporting the use of a
single common pre-processing strategy across all three use cases.

\begin{table}[t]
  \centering\small
  \begin{minipage}[t]{0.5\textwidth}
    \centering
    \caption{Wildfire: floating-point performance of the three input-band
    configurations (reduced dataset). The retained configuration is in
    \textbf{bold}. All values in \%.}
    \label{tab:bands}
    \setlength{\tabcolsep}{2.5pt}
    \begin{tabular}{@{}lcccc@{}}
      \toprule
      Bands & IoU & F1 & Prec. & Recall \\
      \midrule
      Baseline: 13 bands       & 77.87 & 88.13 & 89.14 & 87.15 \\
      B02, B03, B04 (B, G, R)  & 9.49  & 13.68 & 10.49 & 19.67 \\
      \textbf{B02, B11, B12 (B, SWIR1, SWIR2)} & \textbf{78.85} & \textbf{87.16} & \textbf{83.25} & \textbf{91.46} \\
      \bottomrule
    \end{tabular}
  \end{minipage}\hfill
  \begin{minipage}[t]{0.47\textwidth}
    \centering
    \caption{Wildfire: F1-score of the floating-point and quantized models for
    representative input normalizations (full test set, bands B02/B11/B12). The
    retained scheme is in \textbf{bold}. All values in \%.}
    \label{tab:norm}
    \setlength{\tabcolsep}{3pt}
    \begin{tabular}{@{}lccc@{}}
      \toprule
      Normalization & Float & Quant. & $\Delta$(Q$-$F) \\
      \midrule
      No normalization                & 91.42 & 82.32 & $-9.09$ \\
      Scaling $\div128$               & 91.45 & 86.07 & $-5.38$ \\
      \new{\textbf{Scaling $\div$65,535}} & \new{\textbf{95.10}} & \new{\textbf{94.68}} & \new{$\mathbf{-0.42}$} \\
      Per-band centering              & 92.08 & 0.21  & $-91.87$ \\
      $[-1,1]$ after clipping p1/p99  & 83.28 & 81.50 & $-1.78$ \\
      $[-1,1]$ after clipping p5/p95  & 79.77 & 77.86 & \new{$-1.91$}\del{$-1.18$} \\
      \bottomrule
    \end{tabular}
  \end{minipage}
\end{table}

\paragraph{Deployment on the VCK190}
Table~\ref{tab:variants} reports the three variants of the retained
configuration: floating point on the server, quantized on the server, and
quantized deployed on the VCK190. Quantization accounts for essentially all of
the (small) accuracy loss, coming mostly from recall; deployment on the board
adds almost no further degradation. The three variants stay within
${\sim}0.5\,\%$ F1 of one another and their outputs are visually
indistinguishable.
\subsubsection{Hardware Performance on VCK190}
\label{sec:exp-hw}

\paragraph{Inference and end-to-end latency}
On the board, a single DPU inference for a batch of six $1{,}024\times1{,}024$
tiles takes 73\,ms \new{(Table~\ref{tab:latency})}, i.e.\ a throughput of about 86.2\,Mpix/s
($6\times1{,}024^2$ pixels in 73\,ms). The DPU inference is therefore not the
bottleneck: in the un-optimized chain, pre-processing (${\sim}133$\,ms/batch)
and, above all, post-processing (${\sim}256$\,ms/batch) dominate the per-batch
time.

\begin{table}[t]
  \centering\small
  \begin{minipage}[t]{0.42\textwidth}
    \centering
    \caption{Wildfire: algorithmic performance of the three model variants on
    the full test set (bands B02/B11/B12, division by 65,535). All values in
    \%.}
    \label{tab:variants}
    \setlength{\tabcolsep}{4pt}
    \begin{tabular}{@{}lccc@{}}
      \toprule
      Model & Precision & Recall & F1 \\
      \midrule
      Float (server)     & 92.27 & 98.11 & 95.10 \\
      Quantized (server) & 92.50 & 96.96 & 94.68 \\
      Quantized (VCK190) & 92.15 & 97.28 & 94.65 \\
      \bottomrule
    \end{tabular}
  \end{minipage}\hfill
  \begin{minipage}[t]{0.54\textwidth}
    \centering
    \caption{Wildfire: per-stage latency on the VCK190 for a batch of six
    tiles, before and after optimization \new{(mean values)}\del{(mean $\pm$
    std)}.}
    \label{tab:latency}
    \setlength{\tabcolsep}{4pt}
    \begin{tabular}{@{}lccc@{}}
      \toprule
      Stage & Baseline (ms) & Optimized (ms) & Speed-up \\
      \midrule
      Pre-processing  & 132.6 & 31.3  & $\times$4.24 \\
      DPU inference   & 72.7  & 75.2  & $\times$0.967 \\
      Post-processing & 255.9 & 2.4   & $\times$107 \\
      \midrule
      Batch total     & 461.2 & 108.9 & $\times$4.24 \\
      \bottomrule
    \end{tabular}
  \end{minipage}
\end{table}

\paragraph{Optimizing pre- and post-processing}
Both bottlenecks are elementwise conversions that can be precomputed into
Look-Up Tables (LUTs). On the input side, the division by 65,535 and the
subsequent int8 quantization scale are folded into a single 65,535-entry LUT
(64\,KB) that maps every possible uint16 value directly to its int8 code. On the
output side, post-processing consists of a single operation: the DPU output is
dequantized with the output scale and passed through a sigmoid to obtain the
per-pixel fire score; this too is replaced by a LUT. Tabulating these
conversions\new{, together with compiler optimization flags enabling SIMD
vectorization,} greatly accelerates both stages: as shown in
Table~\ref{tab:latency}, pre-processing drops to \new{${\sim}31$\,ms}
and post-processing to ${\sim}2$\,ms per batch, reducing the total per-batch
latency from \new{${\sim}461$\,ms to ${\sim}109$\,ms ($\times$4.2)}, with
bit-exact predictions before and after optimization. The DPU inference stage is
unchanged, as expected.
\new{The optimized chain thus processes ${\sim}58$\,Mpix/s, i.e.\ a
$1{,}024\times6{,}144$ test vector in about 0.11\,s.}

\paragraph{Power consumption}
The board draws about 30\,W on average. Processing adds a modest overhead that
concentrates on the DPU inference stage, which raises consumption by at most
${\sim}6$\,W---peaking around 40\,W during inference---while the pre- and
post-processing stages contribute only a fraction of a watt each. On-board AI
inference on the VCK190 is therefore inexpensive in power relative to the
platform's static budget.

\subsection{\new{Performance at Workflow Level within ORCHIDE}}
\label{sec:exp-orch}

While the previous section describes performances in ideal conditions, with no
sandboxing, concurrency or network latency, this section explores the
performances of the processing chain in the setup described in
Section~\secref{sec:method-orch} to measure orchestration overhead and its causes
and multicore scalability. For reproducibility and due to technical
constraints, these experiments use the LX2160 node as the processing node
(accessing remotely the Versal's DPU during inference). Running the ORCHIDE
runtime itself uses 300m CPU and 4800\,MiB of memory, leaving approximately
11\,GiB and 15.5 cores available for workflows.

As ORCHIDE requires elementary processing steps to be
unikernels~\cite{madhavapeddy2013unikernels}, all processing steps have been
ported to either MirageOS~\cite{madhavapeddy2013unikernels} or
Unikraft\new{~\cite{kuenzer2021unikraft}}. The tiler and mosaicer, being written
in Python which is not easily ported to Unikraft on arm64, have instead been
rewritten (and not optimized) in OCaml as MirageOS \new{unikernels}. On the
other hand, the C++ inference step has not been rewritten but ported to
Unikraft, though the communication with the DPU has been decoupled from the
application using a method based on the vAccel protocol. Compared to
Section~\secref{sec:exp-algo}, all steps are executed as unikernels, i.e.\
virtualized on top of the Linux hypervisor, necessarily introducing further
overhead. The overall time taken to execute ten full processing chains was
measured in four different situations: (1) all chains sequentially without the
orchestrator (i.e.\ running the unikernels directly with no delay), \new{and
(2--4) all chains submitted in parallel to the orchestrator with a varying
limit $L$ on the maximum number of workflows allowed to run simultaneously
($L = 1$, 4 and 10, with $L=1$ equivalent to a fully sequential scheduling)}.
This allows quantifying the overhead incurred by the orchestrator (by comparing
the first two lines) as well as potential parallelization gains when
increasing the concurrency limit close to the number of CPU cores.
All experiments have been carried out on 10 input test vectors of
$1{,}024\times6{,}144$ pixels (i.e.\ ${\sim}6.3$ million pixels). The results
are presented in Table~\ref{tab:orch}. Each line lists the \dely{total} \newy{averaged} clock time
elapsed per workflow (i.e.\ total time for 10 input test vectors divided by 10)
and the ``busy'' time spent actually running steps (as opposed to waiting for
the orchestrator, initializing pods, waiting for disk or network, etc.), as
reported by the orchestrator itself. That total busy time is then
\new{broken down into the share of} each constitutive step of the workflow.
\todo{MFF}{per-step shares computed from the per-step busy times of the Word
table (31/8/75, 5/8/17, 6/8/12\,s), whose totals (114, 30, 26\,s) differ from the
busy times of Fig.~\ref{fig:synthesis}; confirm or replace with the per-step
times of the same runs}
For single workflow operation, we observe that relying on the orchestrator more
than doubles the elapsed total time from input to result \new{(189.6\,s against
78.0\,s)} (due in part to arbitrary waits and pod\newy{s} creation), \new{and still
increases the busy time by about 30\,\% (99.6\,s against 71.3\,s)}\del{even when
comparing busy times only}, which suggests significant overheads due to
Kubernetes pod initialization and runtime. In all cases, we also observe that
the CPU-bound tiler and mosaicer represent the majority of time spent, which
shall drive further optimizations.

\begin{figure}[t]
  \centering
  \begin{subfigure}[b]{0.63\textwidth}
    \centering
    \includegraphics[width=\linewidth]{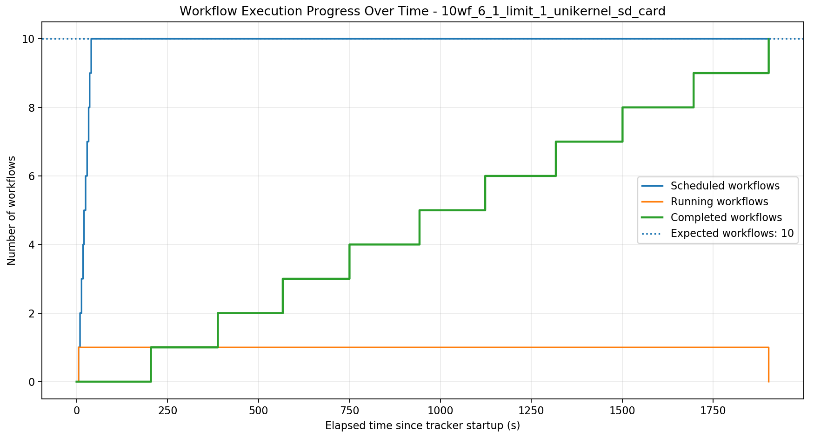}
    \caption{}
    \label{fig:timeline-l1}
  \end{subfigure}\hfill
  \begin{subfigure}[b]{0.34\textwidth}
    \centering
\definecolor{stSched}{HTML}{1F77B4}
\definecolor{stRun}{HTML}{FF7F0E}
\definecolor{stDone}{HTML}{2CA02C}
\begin{tikzpicture}[
  font=\scriptsize,
  box/.style={draw=gray!70, rounded corners=1.5pt, fill=white, align=center,
              minimum width=27mm, minimum height=5mm, inner sep=1.5pt},
  arr/.style={-{Stealth[length=1.6mm]}, semithick, gray!80!black},
  frame/.style={rounded corners=3pt, very thick},
  lab/.style={font=\scriptsize\bfseries, rotate=90, anchor=south, inner sep=1.5pt}
]
  \def\xl{-1.85}\def\xr{1.85}
  \fill[stSched!7] (\xl,-0.75) rectangle (\xr,-1.75);
  \draw[frame, stSched] (\xl,-0.75) rectangle (\xr,-1.75);
  \node[lab, stSched] at (\xl,-1.25) {Scheduled};
  \fill[stRun!8] (\xl,-2.1) rectangle (\xr,-3.75);
  \draw[frame, stRun] (\xl,-2.1) rectangle (\xr,-3.75);
  \node[lab, stRun] at (\xl,-2.925) {Running};
  \fill[stDone!8] (\xl,-4.1) rectangle (\xr,-5.0);
  \draw[frame, stDone] (\xl,-4.1) rectangle (\xr,-5.0);
  \node[lab, stDone] at (\xl,-4.55) {Completed};

  \node[box, fill=gray!10] (ws) at (-0.3,0) {Working-space creation\\(acquisition copy)};
  \node[diamond, draw=gray!70, fill=white, aspect=2.6, inner sep=0.4pt]
        (dec) at (-0.3,-1.25) {$N_{\mathrm{WF}}<L$\,?};
  \node[box] (dep)  at (-0.3,-2.5)  {Pod deployment};
  \node[box] (busy) at (-0.3,-3.35) {\textbf{Busy}: step computation};
  \node[box] (rel)  at (-0.3,-4.55) {Release resources};

  \draw[arr] (ws) -- (-0.3,-0.75);
  \draw[arr] (dec.south) -- node[right, pos=0.5, inner sep=1.5pt] {yes} (-0.3,-2.1);
  \draw[arr] (dec.east) -- (1.2,-1.25) node[below, pos=0.6, inner sep=1.5pt] {no}
             -- (1.2,-0.95) -- (-0.3,-0.95) -- (dec.north);
  \draw[arr] (dep) -- (busy);
  \draw[arr] (busy.east) -- (1.25,-3.35) -- (1.25,-2.5) -- (dep.east);
  \node[font=\tiny, align=center] at (1.57,-2.925) {next\\step};
  \draw[arr] (-0.3,-3.75) -- (-0.3,-4.1);
\end{tikzpicture}
    \caption{}
    \label{fig:states}
  \end{subfigure}
  \caption{\new{(a)} Timeline of data processing for one end to end inference
  workflow at a time with test vector on SD card\del{ (left)}. \new{(b) Workflow
  states monitored by the orchestrator and plotted in (a): \emph{scheduled}
  (waiting for the concurrency limit $L$), \emph{running} (deployment of the
  pod of each step, then busy computation) and \emph{completed}.}\del{Synoptic
  of events for definition of ARGO measured curves (right).}}
  \label{fig:timeline}
\end{figure}

\begin{figure}[t]
  \centering
  \begin{subfigure}[b]{0.57\textwidth}
    \centering
    \includegraphics[width=\linewidth]{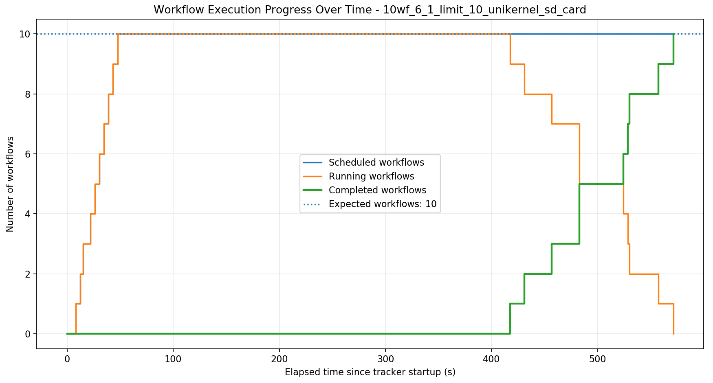}
    \caption{}
    \label{fig:timeline-l10}
  \end{subfigure}\hfill
  \begin{subfigure}[b]{0.4\textwidth}
    \centering
    \includegraphics[width=\linewidth]{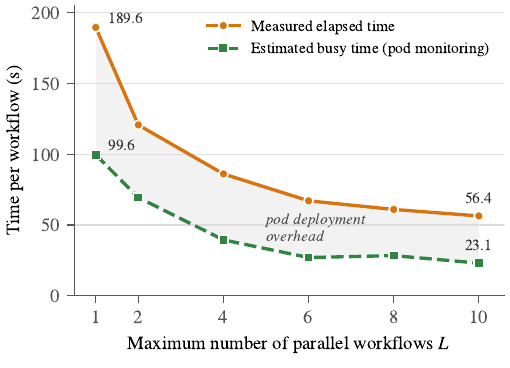}
    \caption{}
    \label{fig:synthesis}
  \end{subfigure}
  \caption{\new{(a)} \new{Timelines}\del{Time lines} for 10 workflows operation in parallel for end
  to end inference\del{ (left)}. \new{(b)} Synthesis of execution times
  depending on the number of workflows in parallel for \del{both tiler and} end
  to end inference\del{ (right)}\new{: measured elapsed time and busy time
  estimated from pod monitoring}.}
  \label{fig:parallel}
\end{figure}

\begin{table}[t]
  \centering\small
  \caption{Time measured or estimated for workflow or its calculation part
  operation and estimation of corresponding throughput\new{, for ten
  $1{,}024\times6{,}144$ test vectors; $L$ is the maximum number of concurrent
  workflows. Elapsed and busy times are those of Fig.~\ref{fig:synthesis}.}}
  \label{tab:orch}
  \setlength{\tabcolsep}{5pt}
  \begin{tabular}{@{}lccccccc@{}}
    \toprule
    & \multicolumn{2}{c}{Total time per WF (s)} & \multicolumn{2}{c}{Throughput (Kpx/s)}
    & \multicolumn{3}{c}{\new{Share of busy time per step (\%)}} \\
    \cmidrule(lr){2-3}\cmidrule(lr){4-5}\cmidrule(l){6-8}
    \new{Configuration} & Elapsed & Busy & Overall & Busy & Tiler & Inference & Mosaicer \\
    \midrule
    No orch., sequential    & 78.0 & 71.3 & 80.7 & 88.3 & \new{22} & \new{5} & \new{73} \\
    Orch., \new{$L=1$}      & \new{189.6} & \new{99.6} & \new{33.2} & \new{63.2} & \new{27} & \new{7}  & \new{66} \\
    Orch., \new{$L=4$}      & \new{86.1}  & \new{39.5} & \new{73.1} & \new{159.4} & \new{17} & \new{27} & \new{57} \\
    Orch., \new{$L=10$}     & \new{56.4}  & \new{23.1} & \new{111.6} & \new{272.8} & \new{23} & \new{31} & \new{46} \\
    \bottomrule
  \end{tabular}
\end{table}
This is illustrated in Fig.~\new{\ref{fig:timeline-l1}}, depicting a timeline
of orchestrator events for one concurrent workflow authorized at a time: a
first delay applies for the creation of a working space associated \new{with the}\del{to}
processed acquisition \newy{(steps in the blue curve)}\new{, then}\del{:} a workflow is scheduled on this
working space \dely{(steps in the blue curve)}. If the maximum limit of \newy{authorized} concurrent
workflows is not reached, scheduled workflows automatically \dely{transition} \newy{turn} to running
workflows (orange curve)\new{,
where each step first deploys its pod and then computes, as summarized in
Fig.~\ref{fig:states}}. From one acquisition processing to another, one
workflow ends to let the next one run. These transitions cannot always be
measured by the orchestrator's built-in monitoring, which allows distinguishing
scheduled, running or completed workflows with a per second resolution. In a
mono-workflow scheme, workflow processing time can be estimated from one rising
edge of one completed workflow to the next (green curve). Since part of these
processing times is spent fetching test vectors from memory, three
configurations have been tested to estimate the impact of the physical support
with respectively test vectors on remote disk, on a local SD card or directly in
a RAM file system. A gap of 9 seconds per workflow has been measured \del{from}
between best and worst cases, suggesting this is \newy{a contributor, but} not the limiting factor. Hence
all results are presented with test vectors on the SD card \newy{of the LX2160 board}.

Fig.~\new{\ref{fig:timeline-l10}}\del{'s left-hand side} depicts the full
timeline of operations when setting the limit of parallel pods to 10.
\new{Fig.~\ref{fig:synthesis}}\del{The right-hand} presents a synthesis of
execution times over a larger range of varying concurrency limits than what is
depicted in Table~\ref{tab:orch}. End to end inference execution time from one
workflow time (\new{189.6\,s}\del{178s} \new{on average,}\del{averaged} obtained from the timeline of
\new{Fig.~\ref{fig:timeline-l1}}\del{the left-hand side of Fig. 4}) is divided
by a factor slightly greater than 3 with \new{56.4\,s}\del{57s} for 10
workflows operating in parallel. Nonlinear behavior of total time with respect
to number of workflows in parallel suggests that \del{that} resource contention
(either CPU, RAM or disk access) quickly limits the gains of parallelization,
though the gains are still significant despite the large overhead of the
orchestrator.

Monitoring at pods level (not detailed here) allows us to distinguish
processing times in the delays from `running' to `completed' events.
\new{Fig.~\ref{fig:synthesis}}\del{The right-hand of Fig. 4} details performed
measurements for various \new{numbers} of workflows authorized at a time. The green
curve corresponds to estimations of end-to-end processing times on the basis of
pods monitoring. The same nonlinear behavior is observed, but the difference
between measured end to end inference workflows operation time curve (orange)
and estimated one from pods monitoring curve (green) suggests that pod
deployment time may contribute from \new{90\,s}\del{77s} ($L=1$, sequential) to
\new{33\,s}\del{40s} ($L = 10$) as inertia in total elapsed times with respect
to time effectively devoted to calculation. \del{A workflow with only the tiler
is represented by the blue curve to isolate its processing time, which
represents 41 to 20s (1 to 10 workflows in parallel).}

\new{\paragraph{Overall latency and throughput}
Put together, the same $1{,}024\times6{,}144$ test vector that the optimized
chain processes in about 0.11\,s on the VCK190 alone
(Section~\secref{sec:exp-hw}) takes 78\,s as a sequence of unikernels without
orchestrator, and between 190\,s ($L=1$) and 56\,s ($L=10$) per workflow under
ORCHIDE, i.e.\ an effective throughput of 33 to 112\,Kpx/s against
${\sim}58$\,Mpix/s. The DPU itself is not the limiting factor: the gap comes
from the un-optimized OCaml tiler and mosaicer, which take most of the busy
time, from the virtualization of every step and the remote access to the DPU,
and from pod deployment, which still accounts for more than half of the elapsed
time at $L=10$. Parallel workflows mitigate but do not remove these overheads,
and the per-workflow latency remains at the minute scale.}

Hence, ORCHIDE workflows instantiation logic shall be challenged in next
developments to recover as \new{much}\del{best} as possible the gap here measured between
performances at algorithm level and the ones at \new{orchestrator} level when
algorithm is associated to a workflow.

\section{Conclusion}
\label{sec:conclusion}

A Generic On-Board AI Processing Chain for Earth Observation has been proposed
in this paper, providing a processing workflow embedding AI applications from
algorithm preparation to mission-level orchestration. Focusing on wildfire
segmentation as the main use case, AI model training and tuning have been
investigated through spectral band selection, input normalization, and int8
quantization. Their impact on algorithmic performance has been evaluated by
comparing floating-point, quantized, and VCK190-deployed models, showing that
the selected configuration enables embedded deployment while preserving model
performance. At hardware level, the execution of the AI processing chain has
also been characterized in terms of latency, throughput, and power consumption,
including the optimization of pre- and post-processing stages. Orchestrator
operation of workflows is then detailed and characterized on an experimental
\new{setup}\del{set up} not representative of final onboard operation but allowing \new{us} to
understand the orchestration mechanism and its dependence on parallel workflow
operation\new{: executed as unikernels under ORCHIDE, the chain benefits from
running several workflows in parallel, which divides the time per workflow by
more than three}.

\new{These results also show the limits of the current set-up: the inference
step accesses the DPU remotely, the tiler and mosaicer unikernels are not
optimized, and pod deployment dominates the elapsed time, so that the latency
per workflow under orchestration remains far from that of the standalone
chain.} These tests at orchestrator level are essential for future mission
planning definition to leverage parallelization by determining the optimal
number of concurrent workflows that can be executed without overloading the
system, ensuring efficient use of available resources\new{. They also}
identify limitations of \del{the} ORCHIDE, such as bottlenecks in pod
allocation or throughput degradations\new{,} and \new{provide the performance data
needed to} optimize mission planning by \del{using performance data to}
fine-tun\new{ing} parameters like acquisition dimensions\new{, i.e.\ the size of
the acquisition clusters processed by one workflow that maximizes throughput
and minimizes processing time. Next steps are to reduce the instantiation
overhead of workflows, to optimize the tiler and mosaicer, to run the whole
chain on the Versal target, and to quantify the data reduction brought by the
downlinked products at mission level.}

\section*{\new{Acknowledgements}}
This work has been supported by French Banque Publique de l'Investissement in
the framework of France 2030. \new{The a}\del{A}uthors would like to thank K\'e\'ephan\new{n} Kokolo
(AKKODIS) for his contribution to these results of characterization\new{, as
well as IRT Saint Exup\'ery and Tarides for their support to this work}.

{\scriptsize
\setlength{\bibsep}{0pt plus 0.1ex}
\linespread{0.93}\selectfont
\begin{multicols}{2}

\end{multicols}
}

\end{document}